%% file: main.tex
\documentclass{article}

\usepackage[preprint]{neurips_2025}

\usepackage[utf8]{inputenc} 
\usepackage[T1]{fontenc}    
\usepackage{hyperref}       
\usepackage{url}            
\usepackage{booktabs}       
\usepackage{amsfonts}       
\usepackage{nicefrac}       
\usepackage{microtype}      
\usepackage[table,xcdraw]{xcolor}
\usepackage{booktabs}
\usepackage{enumitem} 
\usepackage{bbm}
\usepackage{amsmath}
\usepackage{bm}
\usepackage{amsfonts}       
\usepackage{multirow}       
\definecolor{lightblue}{HTML}{7FB2D3}
\definecolor{lightgreen}{HTML}{90B56D}  
\definecolor{darkblue}{HTML}{ec7a2e}  
\definecolor{darkgreen}{HTML}{297645} 
\definecolor{tablepeach}{HTML}{fef2d6}  

\usepackage[normalem]{ulem} 

\usepackage{wrapfig}       
\usepackage{booktabs}   
\usepackage{graphicx}
\usepackage{subcaption}
\hypersetup{
    colorlinks=true,
    breaklinks=true,
    citecolor={green!70!black},
}

\title{RL from Physical Feedback: Aligning Large Motion Models with Humanoid Control}

\author{%
Junpeng Yue$^{1,2}$, Zepeng Wang$^{2,3}$, Yuxuan Wang$^{1,2}$, Weishuai Zeng$^{1}$, Jiangxing Wang$^{1}$
\\ \textbf{Xinrun Xu$^{2}$, Yu Zhang$^{2}$, Sipeng Zheng$^{2}$, Ziluo Ding$^{2}$, Zongqing Lu$^{1,2}$}\thanks{Correspondence to <zongqing.lu@pku.edu.cn>} \\ \\
$^1$PKU \quad $^2$BeingBeyond\quad  $^3$WHU \quad \\ \\
\href{https://beingbeyond.github.io/RLPF/}{https://beingbeyond.github.io/RLPF/}
}

\begin{document}

\maketitle

\input{sec/0_abstract}

\input{sec/1_intro}

\input{sec/2_relatedwork}

\input{sec/3_method}

\input{sec/4_exps}
\input{sec/5_conclusion}


\bibliographystyle{unsrtnat}

\bibliography{neurips_2025}



\appendix

\newpage

\section{Additional Results}

\begin{figure}[ht]
    \centering
    \includegraphics[width=\textwidth]{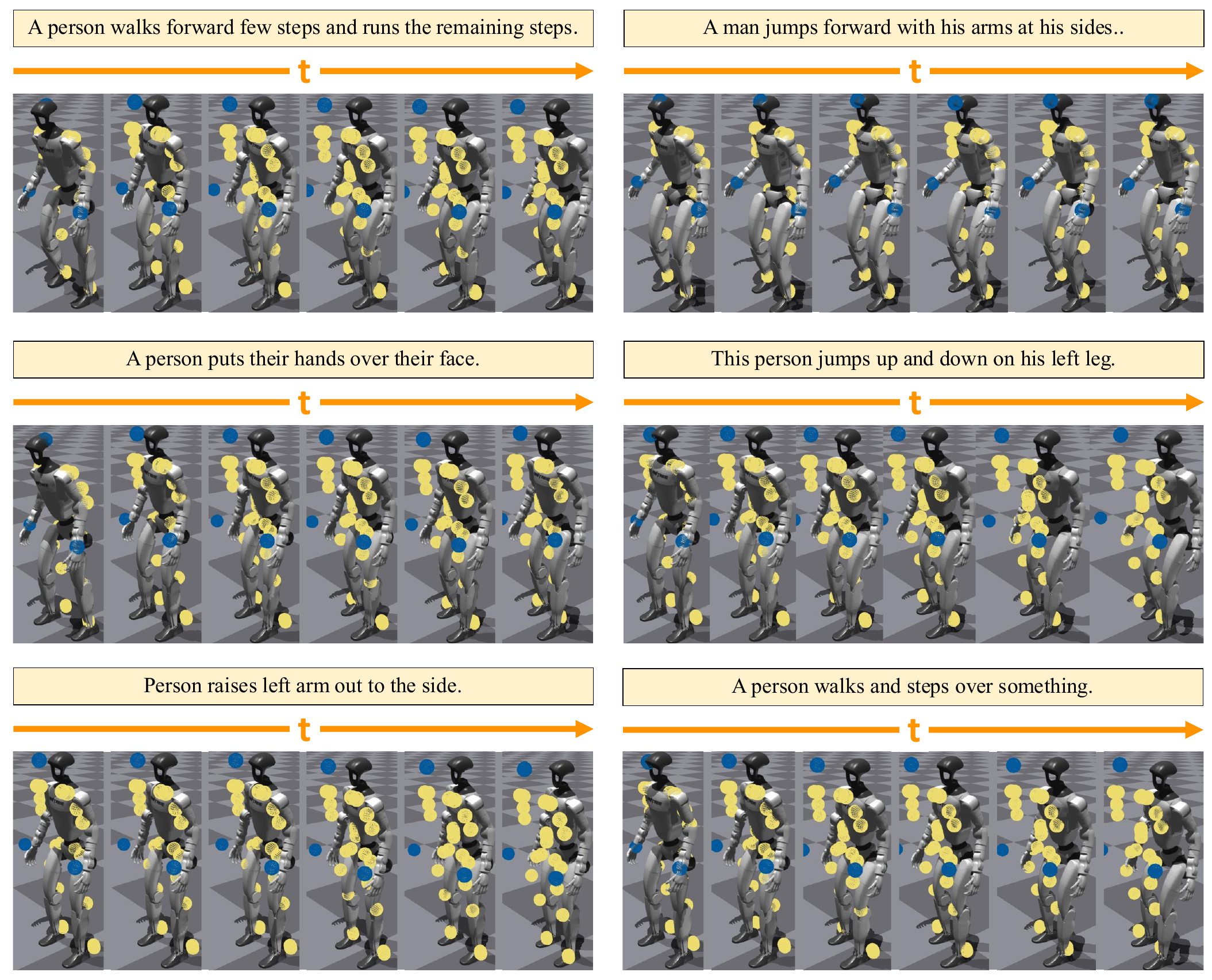}
    \caption{Visualizations of RLPF-w/o align. Since training relies solely on the motion tracking reward, the model predominantly generates standing motion sequences, losing semantic alignment with the input textual instructions.}
    \label{fig:only_stand}
\end{figure}

\subsection{Additional Video Results}

The results of more comprehensive and clear comparisons, particularly real-world videos, presented in accordance with anonymity requirements, are available at \href{https://beingbeyond.github.io/RLPF/}{https://beingbeyond.github.io/RLPF/}.

\subsection{Visualizations of RLPF-w/o align}

As discussed in line 298 of the main text, although\textbf{ RLPF-w/o align} performs very well in low-level evaluations, it nearly completely loses semantic correspondence. In high-level generation evaluations, all metrics exhibit very poor performance.
Visualizations are illustrated in Figure~\ref{fig:only_stand}.

\section{Evaluation Details}

\subsection{Simulator Details}

\begin{figure}[ht]
  \centering
  \begin{subfigure}{0.24\textwidth}
    \centering
    \includegraphics[width=\linewidth]{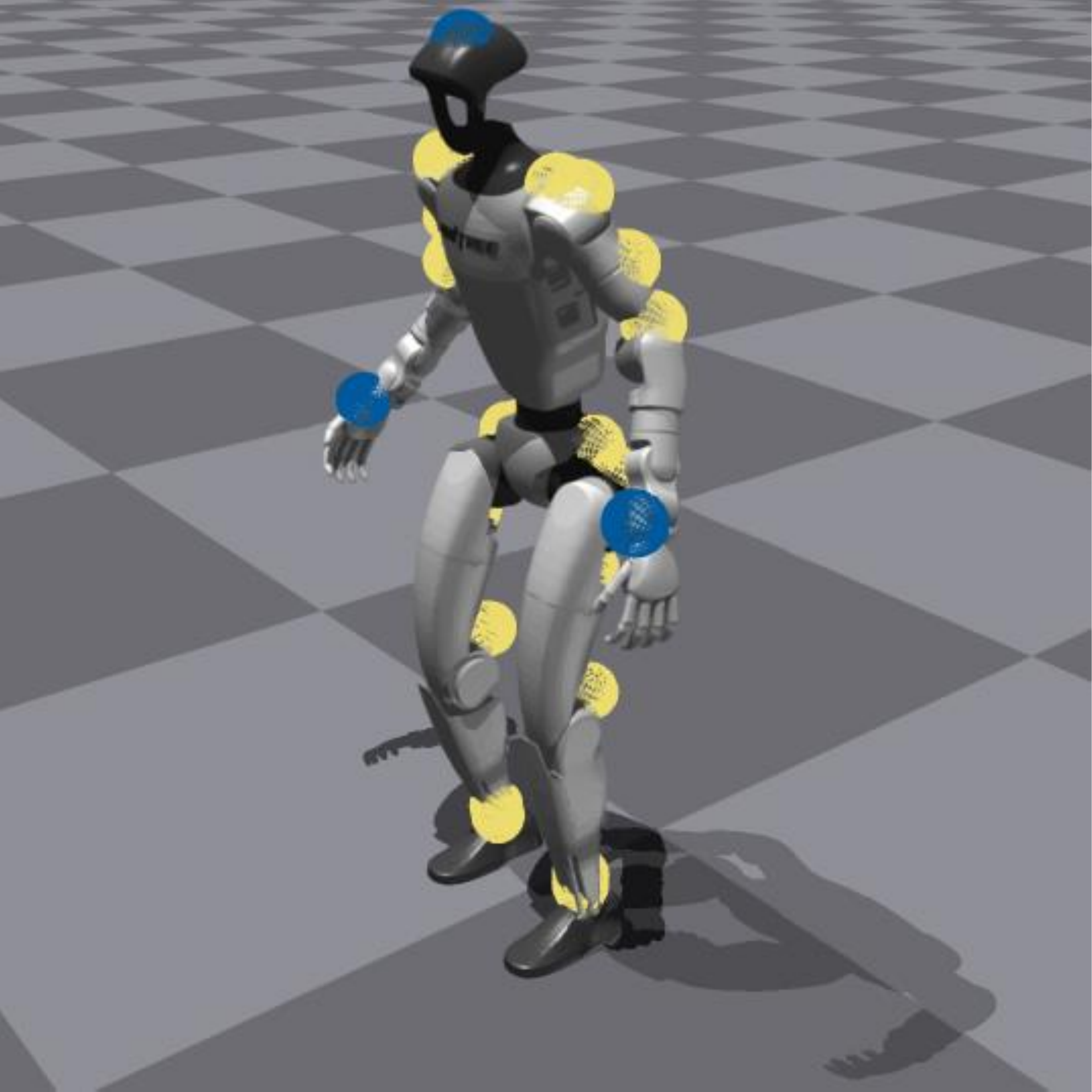}
    \caption{Isaac Gym example 1.}
    \label{subfig:isaacgymex1}
  \end{subfigure}
  \begin{subfigure}{0.24\textwidth}
    \centering
    \includegraphics[width=\linewidth]{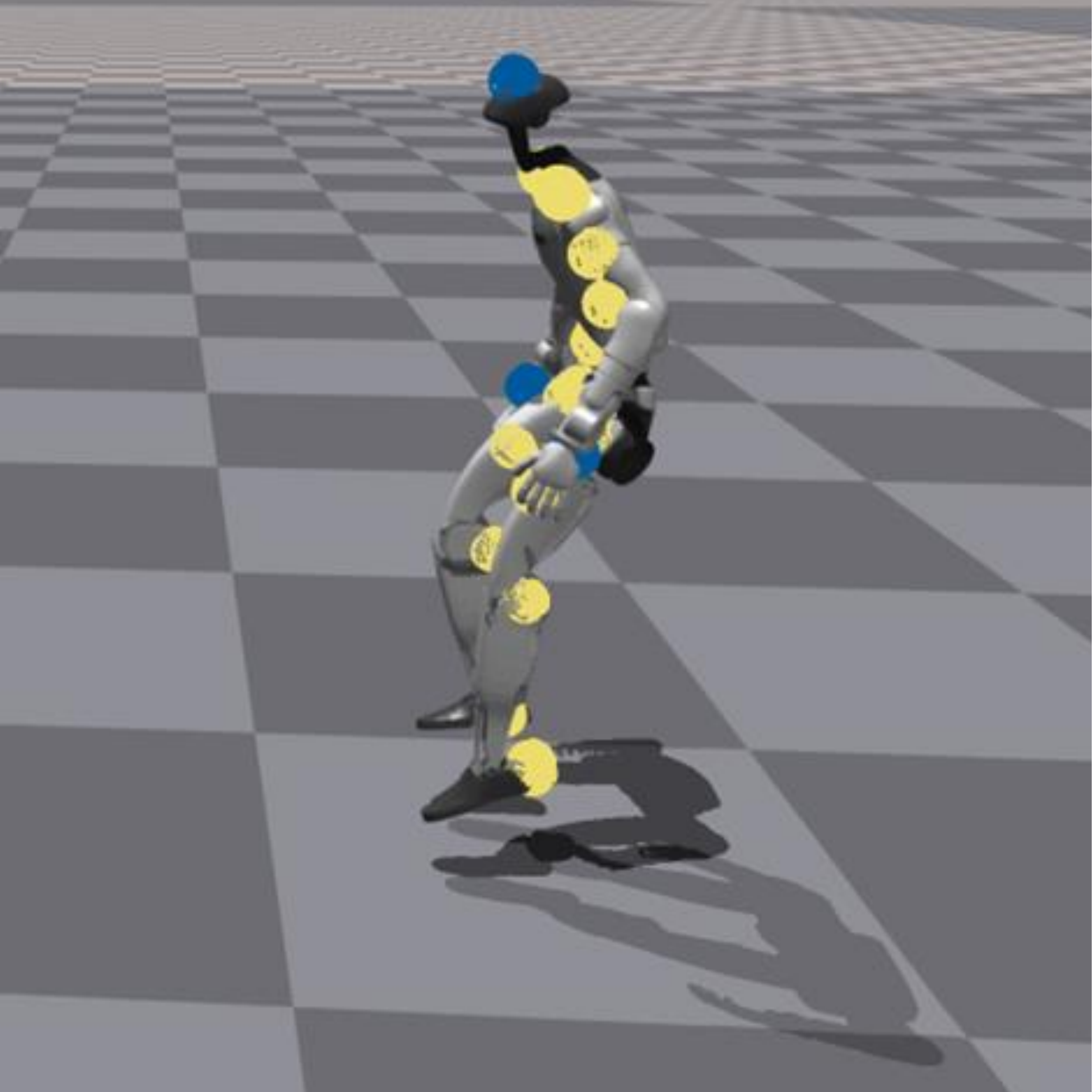}
    \caption{Isaac Gym example 2.}
    \label{subfig:isaacgymex2}
  \end{subfigure}
  \begin{subfigure}{0.24\textwidth}
    \centering
    \includegraphics[width=\linewidth]{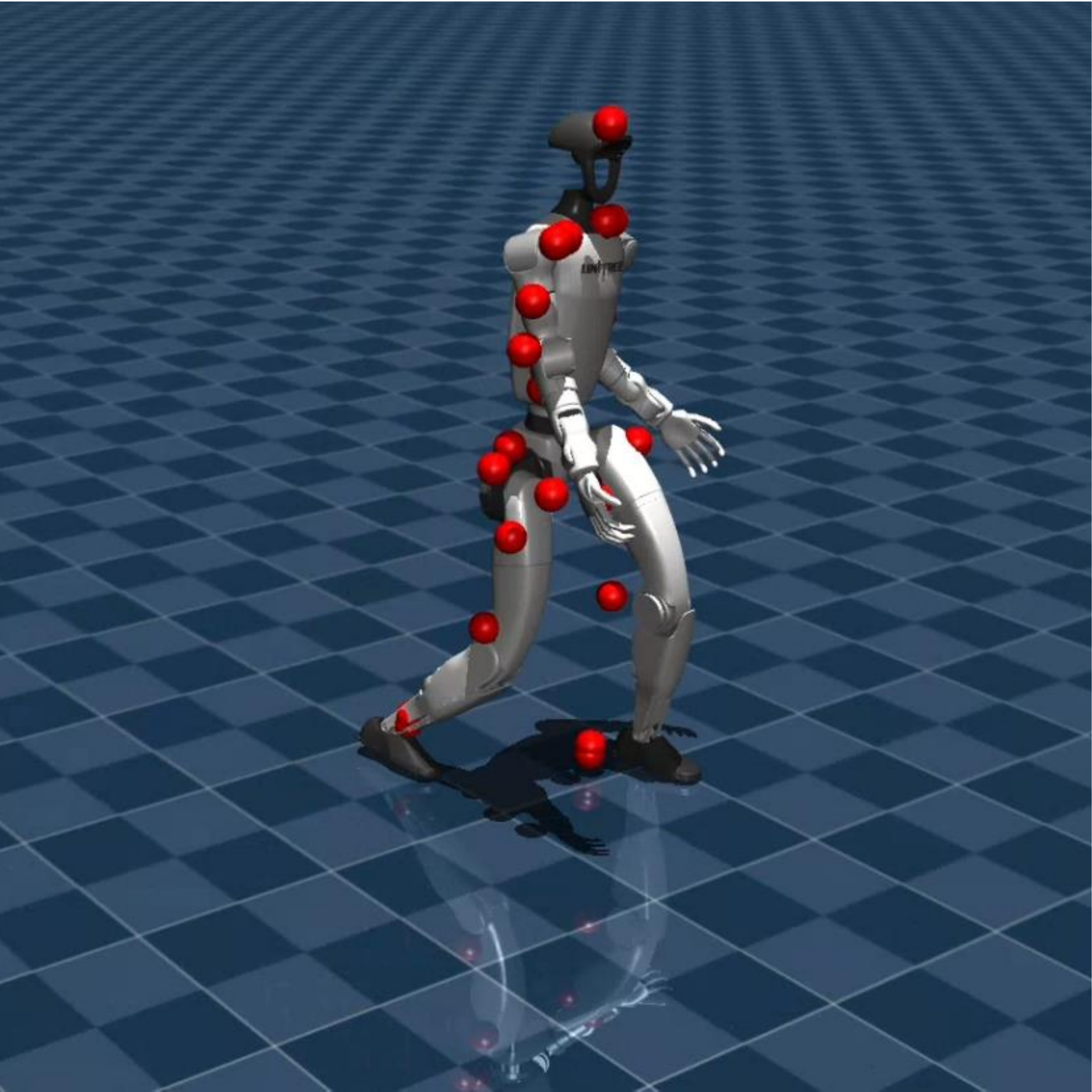}
    \caption{MuJoCo example 1.} 
    \label{subfig:mujocoex1}
  \end{subfigure}
  \begin{subfigure}{0.24\textwidth}
    \centering
    \includegraphics[width=\linewidth]{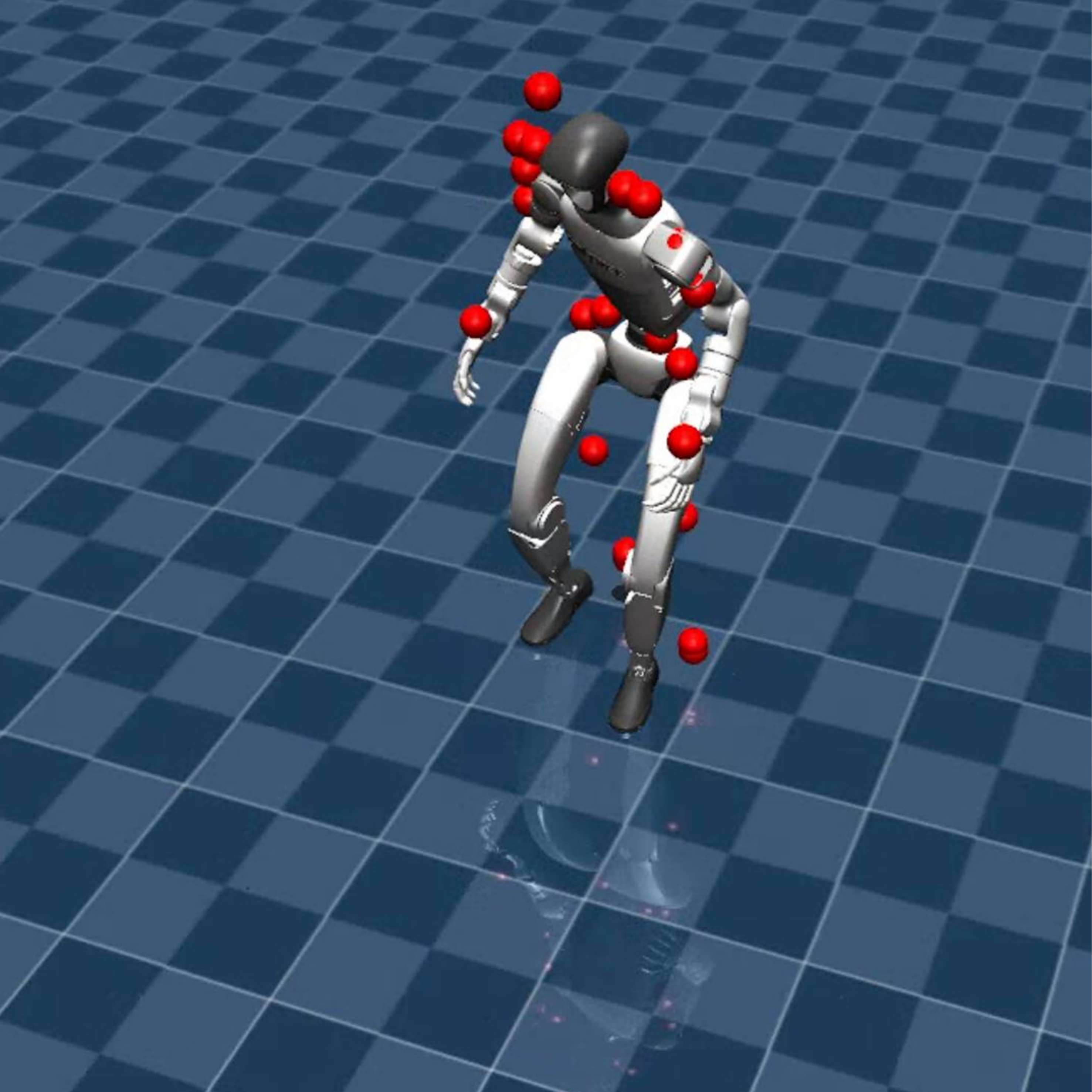}
    \caption{MuJoCo example 2.} 
    \label{subfig:mujocoex2}
  \end{subfigure}

  \caption{Example images of the Isaac Gym and MuJoCo simulators.} 
  \label{fig:env}
\end{figure}

We use the Isaac Gym and MuJoCo simulators to conduct our experiments. The two simulation environments are illustrated in Fig.~\ref{fig:env}.

\subsubsection{Isaac Gym}

Isaac Gym~\citep{makoviychuk2021isaac} is a high-performance physics simulation environment developed by NVIDIA, designed for robotic reinforcement learning (RL) tasks. It uses GPU acceleration to implement an end-to-end training process, significantly improving the simulation and training speed of complex robotic tasks. Isaac Gym supports importing standard robot model formats (such as URDF and MJCF), provides Tensor APIs for direct GPU interaction, and can simulate multiple environment instances in parallel, which is critical for reinforcement learning.

\subsubsection{MuJoCo}

MuJoCo (Multi-Joint dynamics with Contact)~\citep{todorov2012mujoco} is an open source physics engine designed for simulating complex physical systems involving multi-joint structures and contact interactions. It is widely used in the fields of robotics, biomechanics, and machine learning (especially reinforcement learning). MuJoCo is known for its unique combination of speed, accuracy, and modeling capabilities, and is particularly suitable for model-based optimization and contact-rich simulation scenarios.

\subsubsection{Simulator Transfer}

As shown in Humanoid-Gym~\citep{gu2024humanoid}, the dynamics in MuJoCo are closer to the real environment than Isaac Gym. Therefore, following previous work on motion tracking policies~\citep{ji2024exbody2,he2025asap}, we first conduct large-scale reinforcement learning training in IsaacGym. To evaluate the robustness and generalization of the policy, we then perform a zero-shot transfer to the MuJoCo simulator to verify its transferability. 
Finally, we deploy the policy and generated motions on a humanoid robot in the real world to verify the effectiveness of RLPF for the motion tracking task.

\subsection{Evaluation Metrics Details}

\subsubsection{Low-level Tracking Metrics}

For the motion tracking process, we follow OmniH2O~\citep{ji2024exbody2} and PHC~\citep{luo2023perpetual}, and adopt the following metrics: Success Rate, MPJPE, and MPKPE.
\begin{itemize}
    \item \textbf{Success Rate}, denoted as $Succ$, measures whether the humanoid robot can follow the reference motion without losing balance. Tracking is deemed unsuccessful if the average deviation from the reference trajectory exceeds 0.5 meters at any time point, or if the root pitch angle surpasses a specified threshold.
    \item \textbf{Mean Per Joint Position Error (MPJPE)}, denoted as $E_{mpjpe}$ (rad), evaluates the accuracy of joint tracking. It quantifies the average error in degrees of freedom (DoF) rotations between the reference motion and the tracking trajectory.
    \item \textbf{Mean Per Keybody Position Error (MPKPE)}, denoted as $E_{mpkpe}$ (m), evaluates the accuracy of keypoint position tracking. It quantifies the average positional error between keypoints in the reference motion and those in the tracking trajectory.
\end{itemize}

\subsubsection{High-level Generation Metrics}

Following \citet{guo2022generating}, we adopt text-motion retrieval metrics (R@1, R@2, R@3), Multimodal Distance (MMDist), and Fréchet Inception Distance (FID).

\begin{itemize}
    \item \textbf{Text-Motion Retrieval Metrics (R@1, R@2, R@3)}. These metrics assess the quality of text-to-motion generation by evaluating how accurately the generated motions correspond to input text instructions in a retrieval setting. R@1 (Recall at 1) represents the proportion of cases in which the correct motion is ranked as the top match for a given text query. It reflects the model's capability to retrieve the most relevant motion. R@2 and R@3 follow the same principle, indicating the frequency with which the correct motion appears within the top 2 and top 3 retrieved results, respectively.
    \item \textbf{Multimodal Distance (MMDist)}. MMDist measures the average distance between the feature representations of the generated motion and the corresponding text instruction in a shared embedding space. Typically, embeddings are obtained using pretrained text-motion-retrieval models, and a distance metric is then applied. A lower MMDist indicates better alignment between the text and motion, suggesting that the generated motion closely matches the semantic content of the text.
    \item \textbf{Fréchet Inception Distance (FID)}. FID is a widely used metric to evaluate the quality of generated motion by comparing the distribution of generated motions to that of ground-truth motions. It computes the Fréchet distance between two multivariate Gaussian distributions modeled on the feature representations of ground-truth and generated motions, typically obtained using a pretrained motion feature extractor (e.g., an Inception-like network). A lower FID score signifies that the generated motions more closely resemble real motions in both quality and distribution, indicating higher realism and fidelity.
\end{itemize}

\section{Training Details}

\subsection{Dataset Details}

To enhance the text-to-motion generation capabilities of the LLM backbone, we pre-trained it on the large-scale MotionX dataset~\citep{lin2023motion}, and the motion tokenizer was trained on the same data. MotionX includes data from high-quality optical motion capture and relatively lower-quality motion estimation models derived from third-person videos.

During RL fine-tuning, to further enhance the physical plausibility of the generated motions, we used fully motion-captured datasets, CMU~\citep{cmu_mocap} and AMASS~\citep{mahmood2019amass}, which are subsets of MotionX. The quantities of the datasets are summarized in Table~\ref{tab:dataset}.

\begin{table}[ht]
\centering
\caption{Comparison of Dataset Quantities}
\label{tab:dataset}
\begin{tabular}{cc}
\toprule
\textbf{Dataset} & \textbf{Motion Sequence} \\
\midrule
MotionX       & 81,082 \\
AMASS        & 13,145 \\
CMU    & 5,458 \\
\bottomrule
\end{tabular}
\end{table}

\subsection{Motion Retargeting Details}

As described in line 154 of the main text, we follow the idea of H2O~\citep{he2024learning} and adopt a two-step optimization-based approach to achieve the retargeting process. 

Since SMPL parameters represent various human body shapes, we first optimize the shape parameter $\beta^\prime$ to approximate a humanoid form. We select 14 body links corresponding between humans and humanoids, as shown in Table~\ref{tab:link_mapping}, and perform gradient descent on $\beta^\prime$ to minimize joint distances in the rest pose. 

Using the optimized shape $\beta^\prime$ along with the original translation $p$ and pose $\theta$, we perform gradient descent to further reduce the distances between corresponding body links. This process ensures accurate motion retargeting and generates robot motion sequences.

\begin{table}[ht]
\centering
\caption{Correspondence between Humanoid and Human Body Links}
\label{tab:link_mapping}
\begin{tabular}{cc}
\toprule
\textbf{Humanoid Links Name} & \textbf{Human Body Links Name} \\
\midrule
Pelvis       & Pelvis \\
Left hip pitch link        & Left hip \\
Left knee link    & Left knee \\
Left ankle roll link   & Left ankle \\
Right hip pitch link        & Right hip \\
Right knee link    & Right knee \\
Right ankle roll link   & Right ankle \\
Left shoulder roll link    & Left shoulder \\
Left elbow link  & Left elbow \\
Left hand link & Left hand \\
Right shoulder roll link    & Right shoulder \\
Right elbow link  & Right elbow \\
Right hand link & Right hand \\
Head link        & Head \\
\bottomrule
\end{tabular}
\end{table}

\subsection{Hyper-parameter Setting and Experimental Computational Resources}

The specific parameter settings in the experiment are shown in Table~\ref{tab:hyperparameters}. For computational resources, we use eight A800 GPUs to conduct our experiments.

\begin{table}[ht]
\centering
\caption{Hyperparameters of Large Motion Models}
\begin{tabular}{cc}
\toprule
\textbf{Hyperparameters} & \textbf{Value} \\ \midrule
Model Version & NousResearch/Llama-2-7b-hf \\ 
Per Device Train Batch Size & 1 \\ 
Gradient Accumulation Steps & 1 \\
Num Generations & 20 \\
Max Prompt Length & 100 \\
Max Completion Length & 100 \\
Max Grad Norm & 0.1 \\
Reward Weight Tracking & 10 \\
Reward Weight Align & 2 \\
Weight KL & 1.0 \\
KL Type & k3 \\
\bottomrule
\end{tabular}
\label{tab:hyperparameters}
\end{table}

\subsection{Embodiment Details}

In this work, we use Unitree G1 humanoid robot to conduct experiments. 
The robot is equipped with an onboard Jetson Orin NX, which serves as the primary computing and communication unit. The control policy takes motion tracking targets as input, computes the desired joint positions for each motor, and transmits commands to the robot’s low-level interface. The control policy runs at an inference frequency of 50 Hz. The low-level interface operates at 200 Hz, ensuring smooth real-time control. Communication between the control policy and the low-level interface is implemented using LCM (Lightweight Communications and Marshalling)~\citep{huang2010lcm}.	




\newpage

\end{document}

%% file: sec/0_abstract.tex
\begin{figure}[ht]
    \centering
    \vspace{-6mm}
    \includegraphics[width=1\linewidth]{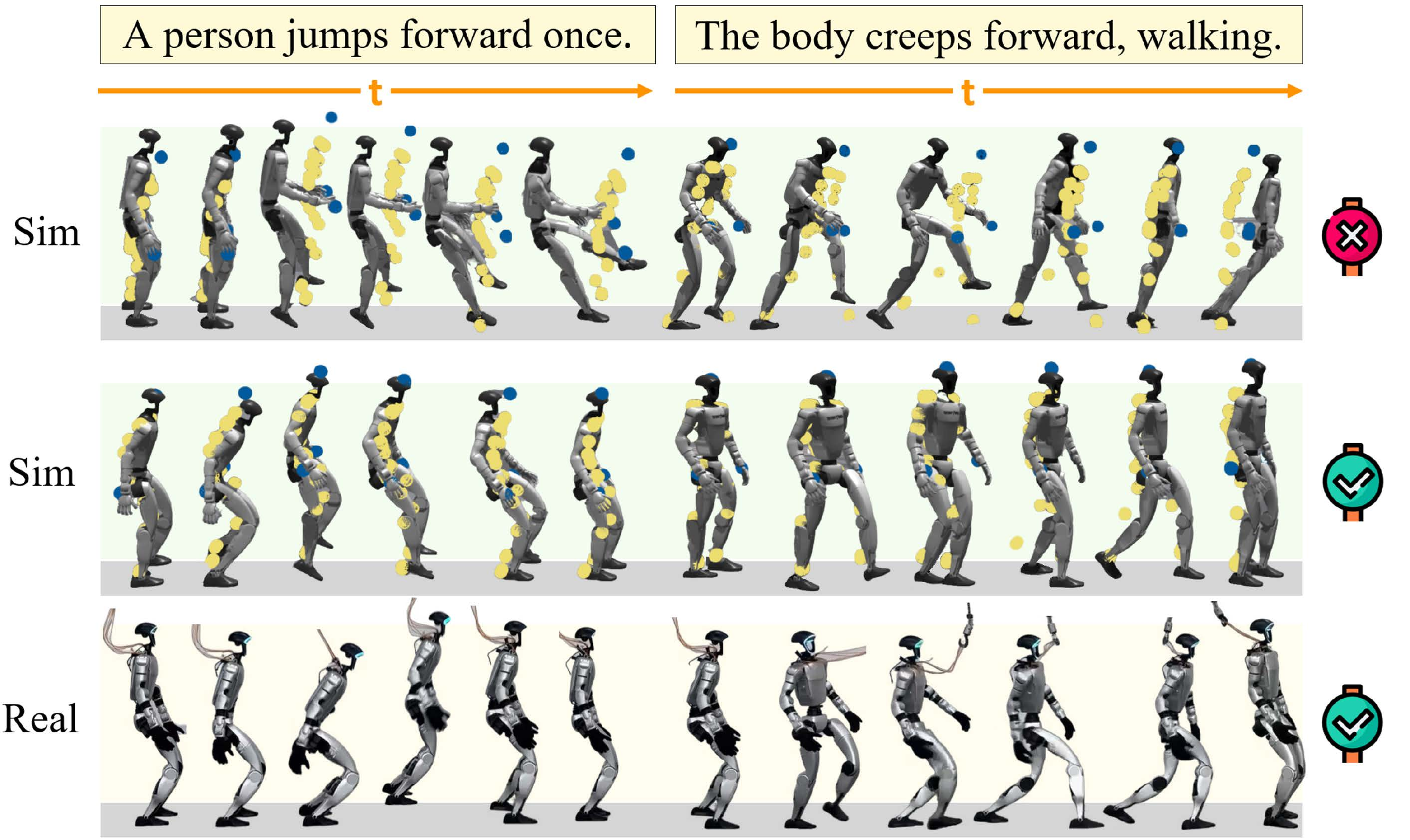}
    \caption{Different behaviors of traditional text-to-motion models and our proposed RLPF in both simulation and real robot.
    While conventional text-to-motion models fail to generate motions conducive to simulation and real machine deployment, RLPF maintains three properties: semantic alignment with descriptions, physical feasibility and practical deployability on robotic systems.}
    \label{fig:first_fig}
\end{figure}

\begin{abstract}
This paper focuses on a critical challenge in robotics: translating text-driven human motions into executable actions for humanoid robots, enabling efficient and cost-effective learning of new behaviors.
While existing text-to-motion generation methods achieve semantic alignment between language and motion, they often produce kinematically or physically infeasible motions unsuitable for real-world deployment. 
To bridge this sim-to-real gap, we propose \textbf{Reinforcement Learning from Physical Feedback (RLPF)}, a novel framework that integrates physics-aware motion evaluation with text-conditioned motion generation.  
RLPF employs a motion tracking policy to assess feasibility in a physics simulator, generating rewards for fine-tuning the motion generator.
Furthermore, RLPF introduces an alignment verification module to preserve semantic fidelity to text instructions.
This joint optimization ensures both physical plausibility and instruction alignment.
Extensive experiments show that RLPF greatly outperforms baseline methods in generating physically feasible motions while maintaining semantic correspondence with text instruction, enabling successful deployment on real humanoid robots.
\end{abstract}

%% file: sec/1_intro.tex
\section{Introduction}
\label{sec:intro}

For decades, the idea of creating lifelike characters that mimic human behavior --- capable of serving as partners in both virtual and physical worlds --- has been a compelling yet elusive vision. 
Recent advances in humanoid robotics (e.g., Unitree G1, Tesla Optimus) have brought this aspiration closer to reality, with hardware achieving greater degrees of freedom (DoF) and reliability.
These improvements enable remarkably fluid motions, such as dancing, that were previously unattainable.
However, this progress is hindered by a critical bottleneck: each motion still requires labor-intensive parameter tuning, severely limiting the scalability of diverse behaviors.

This fundamental limitation has spurred significant interest in automated solutions, particularly text-to-motion (T2M) generation~\cite{lin2023motion}, which formally represents motion as temporal sequences of joint positions and rotations encoding a character's spatial movement.
Existing T2M models demonstrate promising capabilities, generating diverse human motions that provide valuable prior knowledge for humanoid robot control.
Recent integration of large language models (LLMs)~\citep{touvron2023open} has further advanced these systems into large motion models, enhancing both generalizability and human alignment while better interpreting user intent~\citep{wang2024motiongpt,wang2024quo}.
However, a critical gap remains: \textit{Do these state-of-the-art motion models truly fulfill our original vision of enabling humanoid robots to execute the full spectrum of human-like motions with natural fluidity?}

Unfortunately, the answer is \textbf{NO}.
Current motion generation models remain fundamentally limited in their ability to achieve this goal. 
A critical disconnect persists between the motions generated by these models and those executable by humanoid robots, requiring sophisticated motion tracking policies~\citep{ji2024exbody2} to bridge the gap 
between text-guided human motions and physically feasible robot actions. 
This limitation stems from the origins of current human motion research.
Most contemporary text-to-motion generation studies~\citep{zhang2024motiondiffuse, azadi2023make} originate from computer graphics and virtual applications, where physical realism is often secondary to visual quality.  
While these motion generators have achieved remarkable progress in semantic text-motion alignment, they consistently neglect a crucial robotics requirement: \textbf{physical feasibility}. 
The resulting motions frequently violate fundamental physical constraints or ignore dynamic stability considerations --- essential factors for real-world humanoid deployment.
This paper therefore addresses the core research question:

\begin{center}
\textbf{\textit{How can we align motion generation models with whole-body control systems for humanoid robots, ensuring generated motions maintain both semantic accuracy and physical feasibility?
}}
\end{center}

To address this challenge, we must enhance the motion models to generate physically plausible outputs.
Physical feasibility involves two fundamental requirements:
\textbf{\textit{i) Morphological Adaptation}. } 
The inherent morphological differences between humans and humanoid robots make direct motion transfer impractical.
Therefore, effective deployment requires advanced motion retargeting that maintains functional equivalence while strictly adhering to robotic kinematic constraints.
\textbf{\textit{ii) Physical Consistency}. }
Current models frequently produce motions containing non-physical artifacts such as foot sliding, ground penetration~\citep{yuan2023physdiff, han2025reindiffuse, tseng2023edge}, and dynamically unstable body movements.
These violations of physical laws often lead to catastrophic failures in real-world robot execution.
The two challenges highlight the importance of incorporating physical feasibility throughout motion generation for humanoid robot applications.

To tackle these challenges, we propose a novel reinforcement learning (RL) framework for fine-tuning motion generation model for physical feasibility.
Our approach consists of three key components:
\textbf{\textit{i) Motion Tracking Policy. }}
Building upon Exbody2~\citep{ji2024exbody2}, we develop a robust motion tracker that enables precise execution of reference motions while simultaneously providing a signal of physical feasibility.
\textbf{\textit{ii) Alignment Verification Module. }}
We introduce a novel verification mechanism that continuously evaluates the semantic correspondence feedback between generated motions and their textual instructions.
\textbf{\textit{iii) RL-based Finetuning. }}
We employ a reinforcement learning framework that optimizes the motion model using both physical feasibility signals from the tracking policy and semantic alignment feedback from the verification module.

In summary, our key contributions are threefold: 
\begin{itemize}[leftmargin=15pt]
\item We propose \textbf{Reinforcement Learning from Physical Feedback (RLPF)}, a novel framework that optimizes motion generation models through physical feasibility feedback from a motion tracking policy, significantly improving the executability of generated motions on humanoid platforms.
\item We introduce an motion alignment verification module that quantitatively assesses text-motion correspondence, enabling RLPF to maintain semantic fidelity without sacrificing motion diversity or expressiveness.
\item Through extensive experiments, we demonstrate that  our approach achieves state-of-the-art physical feasibility while preserving strong semantic alignment, effectively bridging text-to-motion generation with real-world humanoid control.
\end{itemize}


%% file: sec/2_relatedwork.tex
\section{Related Works}
\label{sec:rw}

\noindent\textbf{Human Motion Generation.}
The rapid development of this task is driven by the availability of high-quality, inscreasing-scale datasets with diverse motions and rich text descriptions.
For instance, AMASS~\citep{mahmood2019amass} unifies 15 motion capture datasets under a standardized framework.
Building on this, HumanML3D~\citep{guo2022generating} offers 14,616 motion sequences and 44,970 text descriptions spanning daily activities, sports, and artistic performances (totaling 28.59 hours).
Further expanding this scale, MotionX~\cite{lin2023motion} introduces 15.6M 3D pose annotations and 81K sequences, leveraging an automated annotation pipeline for efficient and precise data collection.
Additionally, MOVI~\cite{ghorbani2021movi} combines video recordings with real pose and shape data, supporting both generative and discriminative tasks.
These datasets have facilitated the development of advanced text-to-motion models.
T2M-GPT~\citep{zhang2023generating} combines VQ-VAE~\citep{van2017neural} and generative transformers (GPT) to improve alignment with human intent.
MotionGPT~\citep{jiang2023motiongpt} treats human motion as a ``foreign language'' by employing an LLM for unified motion-language modeling.
Its successor, MotionGPT-2~\citep{wang2024motiongpt} extends this framework with multimodal control (e.g., text and pose inputs).
Meanwhile, diffusion-based approaches have achieved high-quality motion synthesis.
MotionDiffuse~\citep{zhang2024motiondiffuse} generates diverse and realistic motions through iterative denoising, while the Motion Diffusion Model (MDM)~\citep{tevet2022human} incorporates geometric losses (e.g., foot contact loss)
from the motion generation domain to achieve state-of-the-art performance.
Our approach leverages these advancements by utilizing pre-trained models to generate initial motion sequences, then fine-tuning their weights via reinforcement learning to produce physically plausible motions for robotics.
Additionally, we optimize motion retargeting by minimizing the distance between human and robot corresponding points while preserving natural movement through regularization.

\noindent\textbf{Reinforcement Learning Fine-Tuning for Large Models.}
RL has emerged as a powerful tool for fine-tuning large pre-trained models, optimizing their performance for specific tasks or aligning them with user preferences.
In natural language processing, reinforcement learning from human feedback (RLHF) leverages human evaluation data --- such as text summarization of tasks~\citep{stiennon2020learning} --- to train reward models and refine outputs using RL algorithms like Proximal Policy Optimization (PPO)~\citep{schulman2017proximal}.
This approach overcomes limitations of traditional supervised fine-tuning for complex objectives.
In human motion generation, RL is increasingly adopted to enhance motion quality, generalization, and real-world applicability. 
For example, InstructMotion~\citep{mao2024learning} frames text-to-motion generation as a Markov decision process, using contrastive pre-trained encoders (text and motion) to design rewards that improve generalization to novel descriptions and reduce overfitting caused by limited training data.
Similarly, RL has been applied to motion control, such as training physics-based tracking policies via PPO for simulated environments~\citep{luo2023perpetual}.
Inspired by these advances, we employs RL to fine-tune motion generation models, prioritizing physical plausibility through physical feedback.
By leveraging pre-trained motion tracking policies to generate rewards, we optimize motions to respect robotic kinematic and dynamic constrains, bridging the gap between generative motion models and practical robotic execution.

%% file: sec/3_method.tex
\section{RL from Physical Feedback (RLPF)}
\label{sec:method}

\begin{figure}[ht]
    \centering
    \includegraphics[width=1\linewidth]{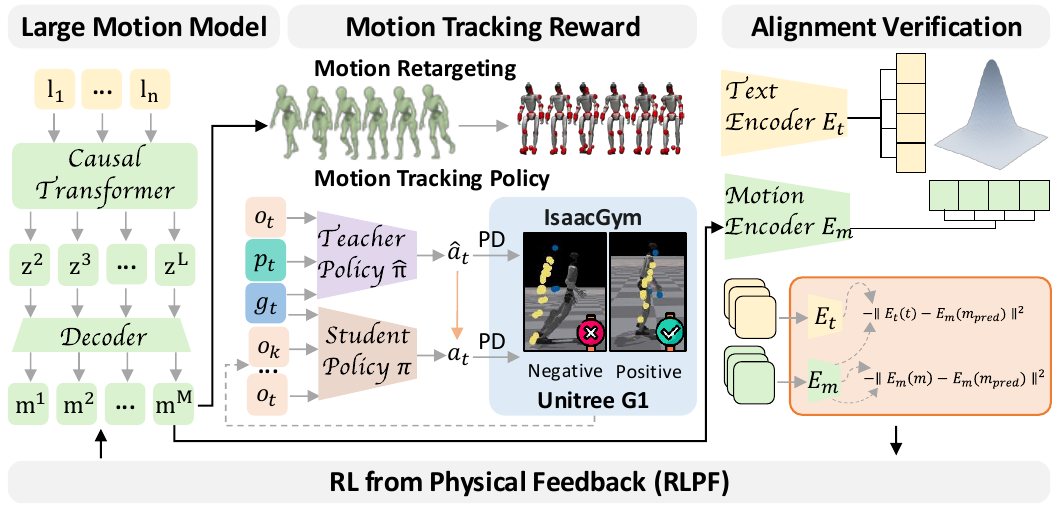}
    \caption{\textbf{Overview of RLPF}, which consists of three key components: \textit{\textbf{i)} Motion Tracking Policy} which is pretrained to establish a motion tracking reward to evaluate generated motions; \textit{\textbf{ii)} Alignment Verification Module} which enhances text-motion semantic consistency while preserving physical plausibility; \textit{\textbf{iii)} RL Optimization Framework} that jointly optimizes the physical feasibility and semantic alignment of motions generated by the large motion model.
    }
    \label{fig:method}

\end{figure}   

This section details key components of our RLPF framework.
First, we establish a foundation for RLPF by pre-training a Text-to-Motion (T2M) generator (Section~\ref{subsec: pretrain}). 
The key innovation of our work, instead, lies in the subsequent reinforcement learning (RL) fine-tuning of this T2M generator (Section~\ref{subsec: RLFT}).
The fine-tuning works based on the guidance of two critical elements: 
\textit{\textbf{i)} a novel motion tracking reward} (Section~\ref{subsec: track}) and \textit{\textbf{ii)} an alignment verification module} to ensure physically plausible and semantically-aligned motions (Section~\ref{subsec: align}).

\subsection{Pre-Training of Text-to-motion (T2M) Generator}
\label{subsec: pretrain}

\noindent\textbf{Motion Tokenizer. }
Existing motion models~\citep{jiang2023motiongpt,wang2024quo} in this field typically employ vector quantization (VQ)~\citep{van2017neural} or its variants~\citep{guo2024momask} to represent human motions, treating them as a ``foreign language''. 
Specifically, these approaches use a motion tokenizer to discretize motion sequences into tokens. The tokenizer comprises three components: \textit{\textbf{i)}} an encoder $\mathcal{E}$ that maps motion sequences to latent codes, \textit{\textbf{ii)}} a decoder $\mathcal{D}$ that reconstructs motions from tokens, and \textit{\textbf{iii)}} a codebook $\mathcal{C}$ where each codeword represents a distinct human body movement pattern.
Given a motion sequence with $M$ frames, the tokenizer first encodes it into motion tokens $\Vec{z}_{1:L} = \{ z_i \}_{i=1}^L$, then decodes them back to the motion $\Vec{m}_{1:M} = \mathcal{D}(\Vec{z}_{1:L}) = \mathcal{D}(\mathcal{E}(\Vec{m}_{1:M}))$.

\noindent\textbf{Large Motion Model for T2M Generation. }
Recent works have shown increasing interest in leveraging large language models (LLMs) for text-to-motion (T2M) generation, capitalizing on their ability to interpret human intent~\cite{jiang2023motiongpt}.
While our approach follows this paradigm, we emphasize that our framework is not restricted to any specific T2M methodology.
For these LLM-based works, the LLM backbone is empowered to process and generate motion features by interesting a motion tokenizer.
In this work, our model extends the LLM's vocabulary with $K$ discrete codes from a motion codebook. 
Our LLM backbone $\Theta$, implemented as a decoder-only causal transformer, first encodes the input text tokens $\mathcal{L}$, then auto-regressively generates corresponding motion tokens $\Vec{z}_{1:L} = \{ z_i \}_{i=1}^L$.
The entire training process of our model consists of two stages: i) large-scale pre-training on text-motion pairs, followed by ii) RL post-training. 
During pre-training, we optimize the model using negative log-likelihood minimization over predicted tokens:
\begin{equation}
\mathcal{L}(\Theta)=-\sum_{j=1}^{T} \log P_{\Theta}(z_j|l, \Vec{z}_{1:j-1}) 
\end{equation}

where $l$ represents the text instruction, $\Vec{z}$ and $z$ denote the generated and target token sequence, respectively.
$\Theta$ represents the model parameters and $T$ is the length of the target sequence.

\subsection{Fine-Tuning of Motion Generator via RL}
\label{subsec: RLFT}

Following pre-training of the base motion generator, we employ RL to fine-tune the model, simultaneously improving both physical plausibility and text-motion alignment.
We formulate the motion generator as an actor policy \( \pi_\theta \), and optimize it using Group Relative Policy Optimization (GRPO).
This RL algorithm offers advantages over standard PPO by eliminating the need for a critic model and instead estimating baselines from relative group scores.
For each text instruction $l$, GRPO operates by: i) sampling a group of motion token sequences $Z = \{\vec{z}^1, \vec{z}^2, \dots, \vec{z}^G\}$ from the previous policy $\pi_{\theta_{old}}$; ii) optimizing the current policy $\pi_{\theta}$ by maximizing the following objective:

{\footnotesize
\begin{equation}
\begin{split}
    \mathcal{J}_{GRPO}(\theta) &= \mathbb{E}{[l \sim P(L), \{\vec{z}^i\}_{i=1}^G \sim \pi_{\theta_{old}}(Z|l)]}  \\
    & \frac{1}{G}\sum_{i=1}^G \left( \min \left( \frac{\pi_\theta(\vec{z}^i|l)}{\pi_{\theta_{old}}(\vec{z}^i|l)} A_i, \text{clip} \left( \frac{\pi_\theta(\vec{z}^i|l)}{\pi_{\theta_{old}}(\vec{z}^i|l)}, 1 - \epsilon, 1 + \epsilon \right)  A_i \right) - \beta \mathbb{D}_{KL}\left(\pi_{\theta} || \pi_{ref}\right)\right)
\end{split} 
\label{eq:GRPO-obj}
\end{equation}
}
\begin{equation}
\mathbb{D}_{KL}\left(\pi_{\theta} || \pi_{ref}\right) = \frac{\pi_{ref}(\vec{m}^i|l)}{\pi_{\theta}(\vec{m}^i|l)}- \log\frac{\pi_{ref}(\vec{m}^i|l)}{\pi_{\theta}(\vec{m}^i|l)} - 1 
\end{equation}

The optimization involves two key hyperparameters: $\epsilon$ controlling the policy update constraint, and $\beta$ balancing the KL-divergence penalty.
Here, $\pi_{\theta_{ref}}$ denotes the reference policy, while $A_i$ represents the advantage computed from a group of rewards $\{r_1, r_2, \ldots, r_G\}$ across all generated outputs within the group:

\begin{equation}
    A_i = \frac{r_i - {\mathrm{mean}(\{r_1, r_2, \cdots, r_G\})}}{{\mathrm{std}(\{r_1, r_2, \cdots, r_G\})}} 
\end{equation}

However, the success of RL fine-tuning still relies on two critical innovation: the motion tracking reward and alignment verification module.
We will detail them in the following sections.

\subsection{Motion Tracking Reward}
\label{subsec: track}
To enable motion deployment on a humanoid robot in simulation for tracking reward computation, we implement a three-stage pipeline: 
\textit{\textbf{i)}} motion retargeting, \textit{\textbf{ii)}} motion tracking and \textit{\textbf{iii)}} offline evaluation.

\subsubsection{Motion Retargeting}

Due to the morphological differences between humans and humanoid robots, human motion must be adjusted to ensure compatibility. 
Following the approach of H2O~\citep{he2024learning}, we employ an optimization-based method for motion retargeting.
Specifically, we first optimize the body shape parameters $\beta$ of the SMPL model~\citep{loper2023smpl} via gradient descent to align its geometric structure with that of the robot, establishing a corresponding human body shape.
With the body shape fixed, we then iteratively adjust the pose parameters $\theta$ by combining inverse kinematics with gradient optimization, ensuring that the joint rotations of the human model match the robot's target posture.
Finally, full-body posture alignment is achieved through keypoint position constraints.
This hierarchical optimization strategy --- decoupling shape adaptation and pose transfer --- ensures physically plausible motion retargeting.

\subsubsection{Motion Tracking Policy}
\label{sec:motion_tracking}

After retargeting, the motion is morphologically compatible with the humanoid robot.
However, it still requires a well-trained motion tracking policy so that the motion can be deployed in simulation and transferred to the real world to evaluate tracking performance.
Inspired by Exbody2~\citep{ji2024exbody2}, we employ a two-stage teacher-student training framework to learn a general motion tracking policy from the AMASS~\citep{mahmood2019amass} dataset. 
First, we train an oracle teacher policy using the off-the-shelf RL algorithm PPO~\citep{schulman2017proximal}, leveraging privileged simulator-only information. 
Then, we distill this policy into a deployable student version by removing privileged information and inaccessible real-world observation. 
All training is conducted in IsaacGym~\citep{makoviychuk2021isaac}, taking advantage of its high-performance parallel simulation.
The details of our policy training procedure is as follows:

\textbf{Stage 1: Teacher Policy Training. }
The teacher policy $\hat{\pi}$ is trained as an oracle in simulation using PPO, leveraging privileged information $p_t$ unavailable in the real world.
This includes ground-truth root velocity, global joint positions, and environmental physical properties (e.g., friction coefficients, motor strength), combined with proprioceptive states $o_t$ and motion tracking targets $g_t$. 
The objective of the teacher policy is to output target joint positions $\hat{a}_t \in \mathbb{R}^{23}$ for Proportional Derivative (PD) controllers, maximizing accumulated rewards to achieve accurate tracking motion tracking and robust behavior.
This yields a high-performing expert policy trained exclusively on simulation data.

\textbf{Stage 2: Student Policy Training. } 
The student policy $\pi$ distills the teacher's knowledge while operating solely on real-world-observable inputs: a history of observations $o_{t-H:t}$ and motion target $g_t$. 
Following a DAgger-like~\citep{ross2011reduction} approach, we first roll out the student in simulation to collect states.
Then we ask the teacher $\hat{\pi}$ to compute oracle actions $\hat{a}_t$ at these states, serving as the supervision signal.
The student is refined by minimizing an MSE loss $l = \Vert a_t - \hat{a}_t \Vert^2$ on aggregated data.
Key to success is maintaining sufficiently long observation histories for effective imitation.
The process iterates until convergence, producing a deployable policy without privileged information.

\subsubsection{Offline Evaluation}

After obtaining a generalizable student policy, we conduct rigorous evaluation in the training simulator (e.g., IsaacGym). 
The policy's tracking performance is quantified using task-specific success rates, which are then used as motion tracking rewards to optimize the motion generator through iterative refinement. 
This process is formalized as:

\begin{equation}
R_{tracking}^{m_i}=\mathbb{I}\left( Succ(\pi, m_i) \right)  
\end{equation}

where $\pi$ denotes the pre-trained student as introduced in Section \ref{sec:motion_tracking}, $m_i$ represents the $i$-th retargeted motion, and $Succ(\pi, m_i) $ is a binary success flag to indicate whether the policy $\pi$ successfully completes the motion tracking task for $m_i$. 
Success criteria incorporate motion quality metrics, with failure conditions defined as: positional deviation exceeding threshold $\epsilon$ or loss of body balance.

\subsection{Motion Alignment Verification}
\label{subsec: align}

We perform RL optimization using the pre-trained motion tracking policy described in Section \ref{sec:motion_tracking}, aiming to generate motion sequences that are easier for the policy to track. 
However, this single-objective optimization may 
compromise the semantic alignment with textual instructions. 
To maintain both semantic fidelity and physical plausibility, we propose an alignment verification module to regularize the motion generator using contrastive-based pretrained encoders ($E_m$ for motion and $E_t$ for text).
Both encoders are trained using the following objective:

\begin{equation}
 \mathcal{L}_{CL} = (1 - y) (\| \mathbf{f_t} - \mathbf{f_m}\|)^2 + y (max(0, m - \| \mathbf{f_t} - \mathbf{f_m}\|))^2
\end{equation}

where $y$ is a binary label indicating text-motion pair matching, $m$ is a margin threshold, and $\mathbf{f_t} = E_t(\mathbf{t})$, $\mathbf{f_m} = E_m(\mathbf{m})$ the respective embeddings.  
The encoders project both modalities into a shared space where semantic similarity corresponds to smaller Euclidean distances.
We implement two verification mechanisms: 
\textit{\textbf{i)}} the \textit{motion alignment verification} which measures deviation from ground truth, and \textit{\textbf{ii)}} the \textit{text alignment verification} which measure fidelity to input description.

\begin{equation}
R_{TA}^{m_i} = (\| \mathbf{E_t(\mathbf{t})} - \mathbf{E_m(\mathbf{m_{\rm pred}})}\|)^2, \,\, R_{MA}^{m_i} = (\| \mathbf{E_m(\mathbf{m})} - \mathbf{E_m(\mathbf{m_{\rm pred}})}\|)^2
\end{equation}

where $R_{TA}^{m_i}$ and $R_{MA}^{m_i}$ represent the text and motion alignment verification, respectively, $m_{\rm pred}$ denotes the generated motion.

%% file: sec/4_exps.tex
\section{Experiments}
\label{sec:exp}

In this section, we conduct a comprehensive evaluation of our proposed method, organized as follows:
First, we describe the experimental setup and baseline comparisons.
We then analyze three key research questions:

\begin{itemize}[leftmargin=15pt]
\item\textbf{Q1 (Feasibility Improvement)}: Does RL fine-tuning enable the motion generator to produce more feasible motions than supervised fine-tuning (SFT), even when SFT uses carefully curated data with tracking-optimized motions?
\item\textbf{Q2 (Reward Analysis)}: How critical is the motion tracking reward to executability, and what are the performance consequences of modifying or removing this reward mechanism?
\item\textbf{Q3 (Alignment Verification)}: Does our alignment verification effectively preserve both text-motion semantic correspondence and kinematic accuracy relative to ground truth?
\end{itemize}

\subsection{Experimental Setup}
\label{subsec:exp_setup}

\noindent\textbf{Compared Baseline.}
To the best of our knowledge, RLPF is the first attempt that employs reinforcement learning to optimize the motion generator for enhanced feasibility and physical plausibility.
We evaluate RLPF against three baseline approaches:
\textit{\textbf{i)} Base Model} which refers to the pretrained motion generator used as our foundation for fine-tuning.
\textit{\textbf{ii)} SFT} which refers to standard supervised fine-tuning on the full training data.
\textit{\textbf{iii)} SFT-Filter} which refers to supervised fine-tuning only on tracking-optimized data, filtered by a motion tracking policy to retain only highly trackable samples.

\noindent\textbf{Simulator Setting. }
Following previous works on motion tracking policies~\citep{ji2024exbody2,he2025asap}, we perform a three-stage evaluation protocol.
First, we conduct large-scale reinforcement learning training in IsaacGym. 
Second, we perform a zero-shot transfer to the MuJoCo simulator to assess the policy's cross-simulator generalization.
Finally, we carry out physical deployment on a Unitree G1 humanoid robot to verify the real-world effectiveness of RLPF on motion tracking.

\noindent\textbf{Dataset. }
To comprehensively validate the effectiveness of our proposed RLPF, we conduct experiments on datasets including CMU~\citep{cmu_mocap} and AMASS~\citep{mahmood2019amass}. 
We use LLaMA2-7B~\citep{touvron2023llama} as our base large language model.

\noindent\textbf{Evaluation Metrics. }
We consider both high-level motion generation and low-level motion tracking metrics to evaluate our motion generator.
evaluate motion generators using both high-level motion generation metrics and low-level motion tracking metrics.

\begin{itemize}[leftmargin=15pt]
\item \textbf{High-Level Motion Tracking Metrics.} 
These metrics evaluate text-to-motion alignment and the fidelity of generated motions compared to their ground truth.
Following \citet{guo2022generating}, we consider text-motion retrieval metrics (R@1, R@2, R@3), Multimodal Distance (MMDist), and Fréchet Inception Distance (FID).
\item \textbf{Low-Level Motion Generation Metrics.} 
We evaluate motions in IsaacGym and MuJoCo before real-robot deployment, aligning with OmniH2O~\citep{ji2024exbody2} and PHC~\citep{luo2023perpetual}.
Key metrics include: 
\textit{\textbf{i)} Success Rate} ($Succ$), which measures whether the humanoid robot follows the reference motion without losing balance (failure threshold: $>0.5m$ average deviation at any timestep).
\textit{\textbf{ii)} Mean Per Joint Position Error} (MPJPE, $E_{mpjpe}$(rad)), which quantifies measures joint tracking accuracy.
\textit{iii) Mean Per Keybody Position Error} (MPKPE, $E_{mpkpe}$ (m)), which evaluates keypoint positional accuracy.
Notably, $Succ$ is the primary metric, as it depends on both MPJPE and MPKPE and incorporates the body’s pitch angle.
\end{itemize}

\input{tab/high_level_eval}

\subsection{High-level Generation Evaluation}
To validate that our approach maintains high-quality motion generation, we evaluate it using high-level generation metrics.
The experimental results demonstrate that our approach achieves significant improvements in low-level tracking while preserving high-level generation accuracy.

\input{tab/low_level_cmu}
\input{tab/low_level_amass}

\subsection{Low-level Tracking Evaluation}

To answer \textbf{Q1 (Feasibility Improvement)}, we first evaluate RLPF's performance on both the CMU and AMASS benchmark test sets.
As shown in Table \ref{tab:low-cmu} and Table \ref{tab:low-amass}, RLPF consistently outperforms all baseline methods across every evaluation metric in both datasets and simulator, clearly establishing its effectiveness.

\input{tab/low_level_abl_cmu}
\input{tab/low_level_abl_amass}

\input{tab/high_level_abl}

\subsection{Ablation Analysis}

We conduct ablation studies to analyze the contribution of RLFS's key components.

To answer \textbf{Q2 (Reward Analysis)}, we compare three variants:  
\textbf{RLPF-Full} (our complete method), 
\textbf{RLPF-PHC} (replacing Exbody2 with PHC~\citep{luo2023perpetual} for motion tracking), and
\textbf{RLPF-w/o track} (ablated by removing the motion tracking reward).
PHC's keypoint-based tracking policy cannot be directly deployed on real robots due to its reliance on simulated environments.
Thus, to evaluate real-world performance, we transfer motions from RLPF-PHC to an Exbody2-based study policy, leading to a transfer gap.
In Table ~\ref{tab:ablation_low_CMU} ~\ref{tab:ablation_low_AMASS} ~\ref{tab:ablation_high}, experimental results demonstrate that the motion tracking reward is essential for enhancing motions' physical feasibility in real-world environments. 
RLPF-Full achieves improved success rates over both RLPF-PHC and RLPF-w/o track across IsaacGym and MuJoCo simulators.

To answer \textbf{Q3 (Alignment Verification)}, we further compare RLPF-Full with another variant \textbf{RLPF-w/o align}, which removes the alignment verification across both CMU and AMASS benchmarks. 
The results in Table~\ref{tab:ablation_low_CMU} ~\ref{tab:ablation_low_AMASS} ~\ref{tab:ablation_high} show that alignment verification is crucial for maintaining motion generation accuracy. 
It's also important to note that RLPF-w/o exhibits near-complete loss of semantic correspondence, as evidenced by the visualizations in our Appendix.


%% file: tab/high_level_eval.tex
\begin{table}[h]
\centering
\caption{Comparisons of High-level generation evaluation on the CMU and AMASS test sets. RLPF-MA and RLPF-TA denote the motion alignment verification and text alignment verification, respectively, as described in Section~\ref{subsec: align}.}
\vspace{2mm}
\scalebox{0.8}{
\begin{tabular}{lcccccccccc}
\toprule
\multirow{2}{*}{\textbf{Method}} & 
\multicolumn{5}{c}{\textcolor{darkblue}{\textbf{CMU}}} & 
\multicolumn{5}{c}{\textcolor{darkgreen}{\textbf{AMASS}}} \\
\cmidrule(lr){2-6} \cmidrule(lr){7-11}
& \textbf{FID ↓} & \textbf{R@1 ↑} & \textbf{R@2 ↑} & \textbf{R@3 ↑} & \textbf{MMDist ↓} 
& \textbf{FID ↓} & \textbf{R@1 ↑} & \textbf{R@2 ↑} & \textbf{R@3 ↑} & \textbf{MMDist ↓} \\
\midrule
\textbf{Base Model} & \cellcolor{tablepeach}{2.35} & 0.24 & 0.38 & 0.49 & 4.17 & 1.79 & 0.22 & 0.35 & 0.46 & 4.62 \\
\textbf{SFT}        & 3.18 & 0.25 & 0.39 & 0.49 & 4.11 & \cellcolor{tablepeach}{1.37} & 0.28 & 0.45 & 0.55 & 4.14 \\
\textbf{SFT-Filter} & 4.11 & 0.19 & 0.34 & 0.46 & 4.22 & 3.10 & 0.24 & 0.40 & 0.52 & 4.21 \\
\midrule
\textbf{GT}         & --   & 0.29   & 0.46   & 0.58   & 3.54   & --   & 0.37 & 0.58 & 0.67 & 3.23 \\
\midrule
\textbf{RLPF-MA} & 3.61 & \cellcolor{tablepeach}{0.26} & \cellcolor{tablepeach}{0.42} & \cellcolor{tablepeach}{0.56} & \cellcolor{tablepeach}{3.63} & 3.34 & \cellcolor{tablepeach}{0.28} & \cellcolor{tablepeach}{0.46} & \cellcolor{tablepeach}{0.58} & 3.75 \\
\textbf{RLPF-TA} & 8.23 & 0.21 & 0.38 & 0.54 & 3.65 & 4.47 & 0.26 & 0.41 & 0.52 & \cellcolor{tablepeach}{3.70} \\
\bottomrule
\end{tabular}}
\label{tab:main-up}
\end{table}

%% file: tab/low_level_cmu.tex
\begin{table}[ht]
\centering
\caption{Comparisons of low-level tracking evaluation on the CMU test set.}
\label{tab:low-cmu}
\vspace{2mm}
\scalebox{0.9}{
\begin{tabular}{lcccccc}
\toprule
 \multirow{2}{*}{\textbf{Method}} &
\multicolumn{3}{c}{\textcolor{darkblue}{\textbf{IsaacGym}}} & 
\multicolumn{3}{c}{\textcolor{darkgreen}{\textbf{MuJoCo}}} \\
\cmidrule(lr){2-4}  \cmidrule(lr){5-7}
 & {\textbf{$Succ$ ↑}} & 
  {\textbf{$E_{mpjpe}$ ↓}} & 
  {\textbf{$E_{mpkpe}$ ↓}} & 
  {\textbf{$Succ$ ↑}} & 
  {\textbf{$E_{mpjpe}$ ↓}} & 
  {\textbf{$E_{mpkpe}$ ↓}} \\
\midrule
\textbf{Base Model}                & 0.43 & 0.19 & 0.18 & 0.43 & 0.35 & 0.88 \\
\textbf{SFT}           & 0.36 & 0.20 & 0.17 & 0.30 & 0.36 & 1.09 \\
\textbf{SFT-Filter}   & 0.57 & 0.19 & 0.16 & 0.41 & 0.31 & 0.73 \\
\midrule
\textbf{RLPF-MA}           & 0.95 & \cellcolor{tablepeach}{\textcolor{black}{0.18}}& \cellcolor{tablepeach}{\textcolor{black}{0.13}} & 
\cellcolor{tablepeach}{\textcolor{black}{0.75}} & \cellcolor{tablepeach}{\textcolor{black}{0.28}} & \cellcolor{tablepeach}{\textcolor{black}{0.23}} \\
\textbf{RLPF-TA}           & \cellcolor{tablepeach}{\textcolor{black}{0.97}} & 0.20 & 0.14 & 
0.61 & 0.30 & 0.36 \\
\bottomrule
\end{tabular}
}
\label{tab:llama2_results}
\end{table}

%% file: tab/low_level_amass.tex
\begin{table}[ht]
\centering
\caption{Comparisons of low-Level tracking evaluation on the AMASS test set.}
\vspace{2mm}
\label{tab:low-amass}
\scalebox{0.9}{
\begin{tabular}{lcccccc}
\toprule
 \multirow{2}{*}{\textbf{Method}}& 
\multicolumn{3}{c}{\textcolor{darkblue}{\textbf{IsaacGym}}} & 
\multicolumn{3}{c}{\textcolor{darkgreen}{\textbf{MuJoCo}}} \\
\cmidrule(lr){2-4} \cmidrule(lr){5-7}
&{\textbf{$Succ$ ↑}} & 
{\textbf{$E_{mpjpe}$ ↓}} & 
{\textbf{$E_{mpkpe}$ ↓}} & 
{\textbf{$Succ$ ↑}} & 
{\textbf{$E_{mpjpe}$ ↓}} & 
{\textbf{$E_{mpkpe}$ ↓}} \\
\midrule
\textbf{Base Model}                & 0.48 & 0.19 & 0.16 & 0.45 & 0.32 & 0.78 \\
\textbf{SFT}           & 0.40 & 0.20 & 0.17 & 0.40 & 0.34 & 0.84     \\
\textbf{SFT-Filter}   & 0.51 & 0.19 & 0.16 & 0.46 & 0.31 & 0.60     \\
\midrule
\textbf{RLPF-MA}           & \cellcolor{tablepeach}{\textcolor{black}{0.92}} & \cellcolor{tablepeach}{\textcolor{black}{0.18}} & \cellcolor{tablepeach}{\textcolor{black}{0.13}} & 0.63 & \cellcolor{tablepeach}{\textcolor{black}{0.28}} & 0.46 \\
\textbf{RLPF-TA}           & 0.90 & 0.19 & 0.15 & \cellcolor{tablepeach}{\textcolor{black}{0.66}} & \cellcolor{tablepeach}{\textcolor{black}{0.28}} & \cellcolor{tablepeach}{\textcolor{black}{0.28}} \\
\bottomrule
\end{tabular}
}
\label{tab:appendix_results}
\end{table}

%% file: tab/low_level_abl_cmu.tex
\begin{table}[!h]

\centering
\caption{Ablation results of low-level tracking on IsaacGym and MuJoCo simulators (CMU Dataset).}
\vspace{2mm}
\scalebox{0.9}{
\begin{tabular}{lcccccc}
\toprule
\multirow{2}{*}{\textbf{Method}} & \multicolumn{3}{c}{\textcolor{darkblue}{\textbf{IsaacGym}}} & 
\multicolumn{3}{c}{\textcolor{darkgreen}{\textbf{MuJoCo}}} \\
\cmidrule(lr){2-4}  \cmidrule(lr){5-7}
&{\textbf{$Succ$ ↑}} & 
{\textbf{$E_{mpjpe}$ ↓}} & 
{\textbf{$E_{mpkpe}$ ↓}} & 
{\textbf{$Succ$ ↑}} & 
{\textbf{$E_{mpjpe}$ ↓}} & 
{\textbf{$E_{mpkpe}$ ↓}} \\
\midrule
\textbf{RLPF-MA}                 & 0.95 & 0.18 & 0.13 & 0.75 & 0.28 & 0.23 \\
\textbf{RLPF-TA}                 & 0.97 & 0.20 & 0.14 & 0.61 & 0.30 & 0.36 \\
\midrule
\textbf{RLPF-w/o track}      & 0.32 & 0.20 & 0.18 & 0.24 & 0.37 & 1.22 \\
\textbf{RLPF-PHC}      & 0.91 & 0.18 & 0.14 & 0.65& 0.27 & 0.33 \\
\midrule
\textbf{RLPF-w/o align}     & 0.91 & 0.18 & 0.12 & 0.90 & 0.29 & 0.19 \\
\bottomrule
\end{tabular}}
\label{tab:ablation_low_CMU}
\end{table}

%% file: tab/low_level_abl_amass.tex
\begin{table}[!h]

\centering
\caption{Ablation results of low-level tracking on IsaacGym and MuJoCo simulators (AMASS).}
\vspace{2mm}
\scalebox{0.9}{
\begin{tabular}{lcccccc}
\toprule
\multirow{2}{*}{\textbf{Method}} & \multicolumn{3}{c}{\textcolor{darkblue}{\textbf{IsaacGym}}} & 
\multicolumn{3}{c}{\textcolor{darkgreen}{\textbf{MuJoCo}}} \\
\cmidrule(lr){2-4}  \cmidrule(lr){5-7}
&{\textbf{$Succ$ ↑}} & 
{\textbf{$E_{mpjpe}$ ↓}} & 
{\textbf{$E_{mpkpe}$ ↓}} & 
{\textbf{$Succ$ ↑}} & 
{\textbf{$E_{mpjpe}$ ↓}} & 
{\textbf{$E_{mpkpe}$ ↓}} \\
\midrule
\textbf{RLPF-MA}                 & 0.92 & 0.18 & 0.13 & 0.63 & 0.28 & 0.46 \\
\textbf{RLPF-TA}                 & 0.90 & 0.19 & 0.15 & 0.66 & 0.28 & 0.28 \\
\midrule
\textbf{RLPF-w/o track}      & 0.30 & 0.20 & 0.18 & 0.30 & 0.35 & 1.10  \\
\textbf{RLPF-PHC}       & 0.80 & 0.18 & 0.14 & 0.61 & 0.28 & 0.41 \\
\midrule
\textbf{RLPF-w/o align}     & 0.99 & 0.18 & 0.11 & 0.99 & 0.29 & 0.13 \\
\bottomrule
\end{tabular}
}
\label{tab:ablation_low_AMASS}
\end{table}

%% file: tab/high_level_abl.tex
\begin{table}[!ht]
\centering
\caption{Ablation results of high-level generation on the CMU and AMASS test sets.}
\vspace{2mm}
\setlength{\belowcaptionskip}{1.5pt}%
\scalebox{0.77}{
\begin{tabular}{lcccccccccc}
\toprule
\multirow{2}{*}{\textbf{Method}} & 
\multicolumn{5}{c}{\textcolor{darkblue}{\textbf{CMU}}} & 
\multicolumn{5}{c}{\textcolor{darkgreen}{\textbf{AMASS}}} \\
\cmidrule(lr){2-6} \cmidrule(lr){7-11}
& \textbf{FID ↓} & \textbf{R@1 ↑} & \textbf{R@2 ↑} & \textbf{R@3 ↑} & \textbf{MMDist ↓} 
& \textbf{FID ↓} & \textbf{R@1 ↑} & \textbf{R@2 ↑} & \textbf{R@3 ↑} & \textbf{MMDist ↓} \\
\midrule
\textbf{RLPF-MA}           & 3.61 & \cellcolor{tablepeach}{0.26}  & \cellcolor{tablepeach}{0.42}  & \cellcolor{tablepeach}{0.56}  & 3.54  & 3.34 & 0.28  &0.46  & 0.58  & 3.75 \\
\textbf{RLPF-TA}           & 8.23 & 0.21  & 0.38  & 0.54  & 3.65  & 4.47 & 0.26  & 0.42  & 0.53  & 3.70 \\
\midrule
\textbf{RLPF-PHC}       & \cellcolor{tablepeach}{3.43} & 0.21  & 0.38  & 0.53  & \cellcolor{tablepeach}{3.56}  & \cellcolor{tablepeach}{1.84} & \cellcolor{tablepeach}{0.30}  & \cellcolor{tablepeach}{0.49}  & \cellcolor{tablepeach}{0.61}  & \cellcolor{tablepeach}{3.64 }\\
\midrule
\textbf{RLPF-w/o align}    & 32.53 & 0.09 & 0.15  & 0.19  & 7.39  & 41.97 & 0.07  & 0.11  & 0.15  & 7.57 \\
\bottomrule
\end{tabular}}
\label{tab:ablation_high}
\end{table}

%% file: sec/5_conclusion.tex
\section{Conclusion}

In this paper, we present Reinforcement Learning from Physical Feedback (RLPF), a novel framework that resolves physical inconsistency in motion generation models for humanoid robots.
RLPF integrates two key components, a pretrained motion tracking policy and an alignment verification module, which ensures that generated motions are both physically feasible and semantically correspond with the input instructions.
Extensive experiments demonstrate the effectiveness of our RLPF approach, showing a pathway for the humanoid robots in real-world applications.


\textbf{Limitations and Future Work. }
The current framework's generalization is constrained by its frozen motion tracking policy, which is pre-trained on a limited static dataset.
A promising direction for future research would be to jointly train the large motion generation model with an adaptable tracking policy, potentially enabling robust handling of out-of-distribution motions while maintaining physical plausibility.
